\documentclass[letterpaper,twocolumn,10pt]{article}

\usepackage[T1]{fontenc}
\usepackage{newtxtext}
\usepackage{helvet}
\usepackage{courier}

\usepackage[hyphens]{url}
\usepackage{graphicx}
\usepackage{amsmath}
\usepackage{booktabs}
\usepackage{natbib}
\setcitestyle{authoryear,round,aysep={ }}
\makeatletter
\def\section{\@startsection {section}{1}{\z@}{-2.0ex plus -0.5ex minus -.2ex}{3pt plus 2pt minus 1pt}{\Large\bf\centering}}
\def\subsection{\@startsection{subsection}{2}{\z@}{-2.0ex plus -0.5ex minus -.2ex}{3pt plus 2pt minus 1pt}{\large\bf\raggedright}}
\def\subsubsection{\@startsection{subsubsection}{3}{\z@}{-6pt plus -2pt minus -1pt}{-1em}{\normalsize\bf}}
\renewcommand\paragraph{\@startsection{paragraph}{4}{\z@}{-6pt plus -2pt minus -1pt}{-1em}{\normalsize\bf}}
\def\@maketitle{%
  \null
  \begin{center}%
    {\LARGE\bfseries \@title \par}%
    \vskip 1.5em%
    {\large \begin{tabular}[t]{c}\@author\end{tabular}\par}%
  \end{center}%
  \par
  \vskip 1.5em}
\makeatother

\title{Ultra-fast Neural Inference for Stochastic Gaussian Splatting Denoising}

\author{
  Chenxiao Hu$^{1}$\quad Hao Zhang$^{1}$\quad Yanchen Zhang$^{1}$\quad
  Meng Gai$^{1}$\quad Guoping Wang$^{1}$\quad Sheng Li$^{1}$\thanks{Corresponding author.}\\[4pt]
  $^{1}$School of Computer Science, Peking University\\
  \{hineven, lisheng\}@pku.edu.cn
}
\date{}

\begin{document}

\maketitle

\begin{abstract}

Stochastic rendering eliminates the sorting and alpha blending process in Gaussian splatting, at the cost of introducing spatial noise.
Formulating temporal denoising over the pixel stream shared by view-consistent stochastic splatting renderers, we propose a temporal neural denoiser validated on stochastic 2D Gaussian Splatting rendering, combining dual-path exponential moving average accumulation, per-pixel learned trust prediction for history validation, a fixed anisotropic spatial filter and a variance-gated composition with stabilization.
The denoiser suppresses the noise, achieving temporally stable, visually compelling outputs during free camera navigation, all while retaining the sort-free, blend-free rasterization performance.
The combined pipeline retains a PSNR gap to sorted alpha-blending renderers, but the denoiser's overhead stays below the time saved by removing sorting and blending.
\end{abstract}

\section{Introduction}
\label{sec:Intro}

Gaussian splatting represents a scene as anisotropic 3D Gaussians (3DGS)~\citep{kerbl2023} or oriented 2D surfels (2DGS)~\citep{huang20242dgs} and renders it through depth sorting and alpha blending. Sorted compositing delivers noise-free images and high visual fidelity, but its cost grows linearly with scene complexity-more primitives, larger frusta, more output channels-until memory bandwidth and fill rate saturate. Engineering works keep the sort-and-blend structure and optimize it directly, delivering large speedups with an unchanged rendering model~\citep{feng2025flashgs,wang2024adr,hanson2025speedysplat}. Order-independent transparency methods remove the global sort and its view-dependent ordering~\citep{hou2024sortfreegs,hahlbohm2025htgs,du2026mobilegs}. A third group redesigns the primitive itself, gaining more compact scenes or more accurate geometry~\citep{ye2025ges,ye2026dpges}. Our work follows a fourth route, stochastic rendering~\citep{kheradmand2025stochasticsplats,sun2025stochasticrt,Rijsdijk2026GaussianPointSplatting}, which removes sorting and blending altogether.

With stochastic rendering, each fragment is retained with probability equal to its opacity contribution and resolved using a standard depth test. This eliminates global sorting, replaces alpha blending with opaque z-buffer compositing, and enables additional output channels to be decoded from a visibility buffer at minimal cost. The resulting image is an unbiased Monte Carlo estimate of conventional alpha blending.
StochasticSplats~\citep{kheradmand2025stochasticsplats} demonstrates that the estimator matches the alpha-blended image in expectation while running fast. Its remaining problem is Monte Carlo noise: at one sample per pixel, the estimate is far too noisy for practical use, and the generic temporal anti-aliasing (TAA, \cite{karis2014}) or frame averaging used to suppress it is insufficient under free camera navigation.

\begin{figure}[t]
    \centering
    \includegraphics[width=\linewidth]{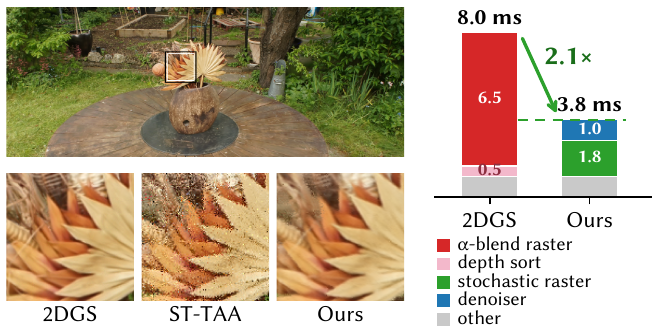}
    \caption{Top left: Garden navigation. Bottom left: zoom-in by conventional $\alpha$-blended 2DGS renderer, by stochastic rasterization with ST-TAA (see Sec. \ref{sec:experiments}), and by our approach on the same setting. ST-TAA suffers from residual Monte Carlo noise, whereas our method visually resembles the conventional renderer. Right: per-frame time breakdown. Despite the additional cost of denoising, our approach achieves significant speedup by avoiding the expensive Gaussian depth sort and $\alpha$-blending required by conventional splatting.}
    \label{fig:motivation}
    \vspace{-0.6cm}
\end{figure}

Temporal accumulation can repair such noise, but it relies on accurate reprojection, which presumes view-consistent geometry. Per-view projection and per-Gaussian sorted compositing violate this assumption. Several works improve the view consistency of splatting by refining the sorting strategy~\citep{radl2024stopthepop,liu2025duplex} or change the primitive or blending model~\citep{shen2024solidgs,huang2024deformable}, and work well within the sorted pipeline; others avoid Gaussian depth sorting altogether, anchoring geometry to world-space surfels whose per-pixel visibility is resolved exactly by z-buffering or depth peeling~\citep{ye2025ges,ye2026dpges}. 
For our experiments, we take a minimal modification of 2DGS from per-Gaussian to per-pixel depth ordering, as a prototype that keeps the scene representation mostly comparable to the vanilla baseline. 

\textbf{Our method.} With semi-transparent fragments jittering between frames and no noise-free G-Buffer guidance, heuristics and generic anti-aliasing methods malfunction; a dedicated denoiser is required. We design a lightweight temporal neural denoiser for view-consistent stochastic splatting. The denoiser maintains two exponential moving averages per pixel: an \textit{accumulated} path accumulates raw stochastic samples and converges to the unbiased alpha-blended image; a \textit{denoised} path accumulates spatially filtered frames, suppressing visually prominent noise in disocclusions with low frame accumulation. A variance gate blends the two averages per pixel. A neural network predicts how much reprojected history to trust per pixel; a fixed anisotropic mipmap spatial kernel with learned per-Gaussian parameters performs the filtering. The pipeline is optimised under a single photometric objective. The denoiser's overhead is far below the time saved by removing sorting and blending, as seen in Fig.~\ref{fig:motivation}.

Our contributions are as follows:
\begin{itemize}
\item A formulation of temporal denoising over the per-pixel stream shared by view-consistent stochastic splatting renderers, and a lightweight neural denoiser for the stream denoising problem, optimised end-to-end and validated on a per-pixel-ordered 2DGS host.
\item An experimental demonstration that this pairing makes stochastic rasterization beneficial: on a minimal per-pixel-ordered 2D Gaussian host, the pipeline approximates the converged Monte-Carlo quality on static views, stays visually satisfying under free navigation, runs faster than the sorted alpha-blending renderer.
\end{itemize}

\section{Related Work}
\subsection{Gaussian Splatting and Rasterization Acceleration}
\label{sec:related:gs}

3D Gaussian Splatting (3DGS)~\citep{kerbl2023} represents a scene as anisotropic 3D Gaussians and renders them with tile-based rasterization: primitives are depth-sorted per frame and alpha-blended front to back, achieving real-time radiance-field rendering with high visual quality.
Keeping the representation and the sort-and-blend model unchanged, a line of work pushes efficiency further through aggressive culling and pipeline scheduling~\citep{feng2025flashgs}, Tensor-Core-friendly reformulation of the per-pixel computations~\citep{li2026tensorgs}, or precise splat localization and pruning~\citep{hanson2025speedysplat}.

\subsection{Gaussian Splatting Variants}
\label{sec:related:variants}

A second body of work modifies the representation or the compositing model itself.
One goal is view-consistency: evaluating each Gaussian at a locally optimal per-pixel depth with hierarchical resorting~\citep{radl2024stopthepop}, replacing alpha blending with an order-independent weighted sum~\citep{hou2024sortfreegs}, anchoring the scene in opaque z-buffered surfels~\citep{ye2025ges}, or recovering the exact per-pixel order of semi-transparent surfels with depth peeling~\citep{ye2026dpges}.
A second goal is efficiency, through adaptive rasterization radii and tile load balancing~\citep{wang2024adr}, order-independent compositing on mobile GPUs~\citep{du2026mobilegs}, or z-buffered opaque layers with selective blending~\citep{hahlbohm2025htgs}.

\paragraph{Stochastic rendering.}
StochasticSplats~\citep{kheradmand2025stochasticsplats} applies stochastic transparency to splats: each fragment is emitted as fully opaque with probability equal to its opacity contribution, which removes sorting and blending and makes each frame an unbiased Monte Carlo estimate of the alpha-blended image.
The simplicity provides performance gain, but also brings severe Monte Carlo noise. 
The output only becomes visually satisfying after accumulating many frames, which assumes a static camera. 
Ray-traced formulations, compatible with stochastic or deterministic sampling~\citep{sun2025stochasticrt, moenneloccoz2024gdgrt,condor2025dontsplat}, sidestep splatting entirely, but forgo the throughput of rasterization.

2D Gaussian Splatting (2DGS)~\citep{huang20242dgs} represents scenes as planar surfels, whose exact ray-splat intersection depth provides more view-consistency.
Our work builds on this property: better consistency makes temporal reprojection more reliable and temporal denoising plausible.

\subsection{Denoising, Super-Resolution, Anti-Aliasing}
\label{sec:related:denoising}

Kernel-predicting denoising, an influential family of Monte Carlo denoisers, predicts filtering kernels for noisy inputs~\citep{bako2017, vogels2018}; related variants regress the clean image from feature-guided inputs~\citep{kalantari2015, bitterli2016}, operate on individual samples~\citep{gharbi2019}, or target self-supervision and interactive budgets~\citep{yu2021mc, back2022, meng2020, isik2021, balint2023}. 
Our denoiser is very similar to this family, but is designed for purely noisy inputs. Regular real-time Monte Carlo denoisers commonly rely on noise-free G-buffers. Stochastic splatting emits no noise-free guidance so existing denoisers cannot be applied to stochastic Gaussian splatting with trivial modification. \cite{zeng2026arbitrary} uses neural network for real-time super-resolution on 3DGS rendering, while we employ neural network for denoising: a different post-processing task.

Temporal Anti-Aliasing~\citep{karis2014} stabilizes frames by reprojecting the history buffer along motion vectors and clipping colors to the local neighborhood. Learning-based supersampling follows the same history-driven paradigm~\citep{xiao2020, zhong2023fusesr, yang2024mnss}.
They expect near-noise-free input and reliable motion vectors, neither of which stochastic splatting provides.

\begin{figure*}
\includegraphics[width=\linewidth]{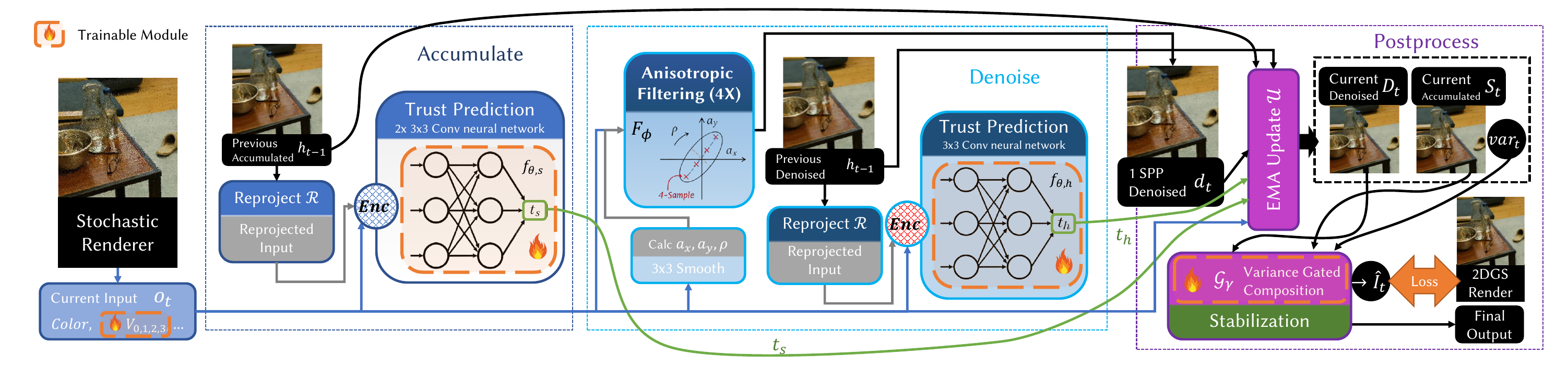}
\vspace{-0.7cm}
\caption{Workflow of our denoiser for each frame (Eq.~\eqref{eq:unified}) with trainable modules highlighted. The host stochastic rasterizer emits the 1\,spp stream (Sec.~\ref{sec:method:stream}). In the \emph{accumulated path} (left), the previous accumulated state is reprojected into the current view and a convolutional predictor estimates the history trust $t_s$ from reprojection-derived motion, depth, and photometric cues. In the \emph{denoised path} (middle), the current frame is spatially filtered by a fixed anisotropic mipmap kernel parameterised by smoothed neural-view scalars, and a lightweight head estimates the trust $t_h$ of the reprojected denoised mean. In \emph{postprocessing} (right), both paths update their exponential moving average (EMA) states (Eq.~\eqref{eq:ema-update}), a variance gate (Eq.~\eqref{eq:gate}) composites the two means per pixel, and a stabilization pass (\emph{STAB}) suppresses residual jitter. The updated states propagate to the next frame. 
}
\label{fig:pipeline}
\end{figure*}

\section{Method}\label{sec:method}
\subsection{Overview}\label{sec:method:overview}
Standard alpha-blended 2DGS produces noise-free images, but pays for per-frame global depth sorting and heavy overdraw from semi-transparent fragments; both costs grow with primitive count, frustum size, and channel count. Stochastic rasterization removes both: each fragment is randomly retained with probability equal to its opacity contribution and resolved by an ordinary depth test, and auxiliary channels decode from the visibility buffer at little extra cost. The resulting image is an unbiased Monte Carlo estimate of the alpha-blended reference, but its single-sample-per-pixel noise is visually prominent.

Temporal filtering is a proven remedy for Monte Carlo noise. The denoiser must cost well below the savings from removing sorting and blending, run without the noise-free G-buffers conventional real-time denoisers rely on, and accumulate history only where the observed surface is viewpoint-invariant: a property standard 2DGS lacks due to per-Gaussian depth ordering.

We therefore formulate temporal denoising over per-pixel stream of view-consistent stochastic renderer (Sec.~\ref{sec:method:stream}) as a recurrent model with sparse learned components:
\begin{equation}
\begin{aligned}
c_t &= \mathcal{R}(o_t, h_{t-1}), & (t_s, t_h) &= f_\theta(c_t),\\
d_t &= F_\phi(o_t), & (S_t, D_t) &= \mathcal{U}(o_t, d_t, h_{t-1}, t_s, t_h),\\
g_t &= \mathcal{G}_\gamma(\mathrm{var}_t, n_s, n_h), & \hat{I}_t &= g_t D_t + (1-g_t) S_t .
\end{aligned}
\label{eq:unified}
\end{equation}
Here $o_t$ is the current stochastic observation (colour, depth, decoded payload), $h_{t-1}$ the temporal state, and $\hat{I}_t$ the output frame. $\mathcal{R}$ reprojects the state into the current view and derives consistency features (Sec.~\ref{sec:impl:reproject}); $f_\theta$ predicts how much reprojected history each of two complementary paths may trust; $F_\phi$ is a fixed anisotropic kernel with footprints parameterised by learned neural-view scalars $\phi$ (Sec.~\ref{sec:impl:spatial}); $\mathcal{U}$ is the exponential-moving-average update (Eq.~\eqref{eq:ema-update}) yielding an accumulated mean $S_t$ over raw samples and a denoised mean $D_t$ over filtered frames; and $\mathcal{G}_\gamma$ is a variance gate (Eq.~\eqref{eq:gate}) fusing the two means per pixel. $\mathcal{R}$, $F$, $\mathcal{U}$, $\mathcal{G}$ are fixed differentiable operators; learning is confined to the trust networks $\theta$, the gate scalar $\gamma$, and the payload $\phi$, which are optimised jointly under a single photometric objective on $\hat{I}_t$ through the same recurrent state transition used at inference (Sec.~\ref{sec:impl:training}). Validation instantiates the simplest host model satisfying the stream requirements (Sec.~\ref{sec:method:model}); Figure~\ref{fig:pipeline} illustrates the full pipeline.

\subsection{The Stochastic Splatting Stream}\label{sec:method:stream}

Per pixel and per frame, a stochastic splatting renderer emits a \emph{visibility sample}: Identity of the primitive winning an opaque depth test after stochastic retention, together with its \emph{depth}; auxiliary per-primitive payload channels (colors, or trainable parameters such as the neural-view scalars in Sec.~\ref{sec:impl:spatial}) are decoded from the visibility sample at little extra cost. Two properties make this stream temporally denoisable: view-consistency, so history can be reprojected safely, and approximately unbiased decoded radiance, so accumulating samples converges to $\alpha$-blended rendering. Noise magnitude and view-consistency vary across such renderers~\citep{kheradmand2025stochasticsplats,sun2025stochasticrt,Rijsdijk2026GaussianPointSplatting}, they have a common issue: Low samples per pixel introduce prominent Monte Carlo noise. This can be effectively suppressed by our denoiser using only the above quantities.

\subsection{Minimal Host Model}\label{sec:method:model}

For validation we instantiate the simplest model that satisfies the requirements: vanilla 2DGS with the per-Gaussian centre-depth ordering replaced by per-pixel depth ordering, rendered with the same stochastic visibility rule as our primary baseline~\citep{kheradmand2025stochasticsplats} so the comparison isolates the denoiser rather than the geometry model. A pretrained vanilla 2DGS scene is converted by full finetuning through a depth-peeling differentiable rasterizer~\citep{everitt2001depthpeeling,laine2020nvdiffrast}; the conversion is not free, leaving a residual gap of about 1\,dB to vanilla 2DGS on average. We deliberately keep the dense, unpruned reconstruction so the geometry stays comparable to the 2DGS baseline.

\subsection{Denoiser Design}\label{sec:method:denoiser}

The compute budget left by the rasterization savings is narrow, and the stream offers no noise-free G-Buffer. Fixed heuristics, e.g. blending by depth difference alone, are therefore insufficient. We adopt a kernel-estimation-style denoiser that spends most of its budget predicting, per pixel, how much reprojected history can be trusted ($f_\theta$ of Eq.~\eqref{eq:unified}), using only stream quantities (Sec.~\ref{sec:method:stream}).

\textbf{Accumulated and Denoised paths.}
\begin{figure}
    \centering
    \includegraphics[width=0.95\linewidth]{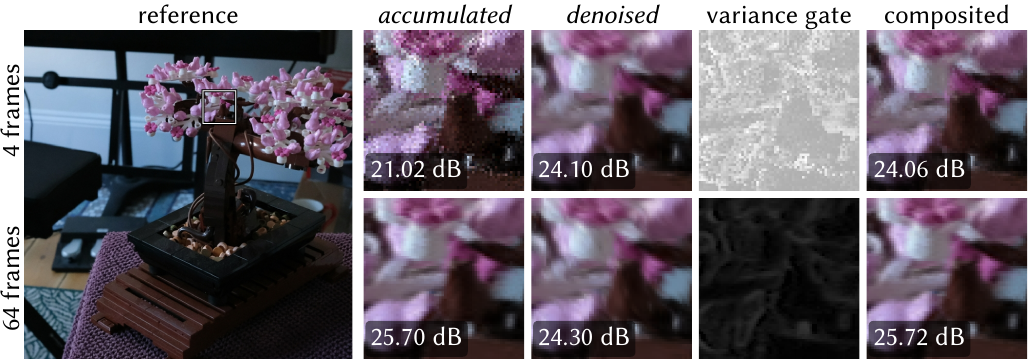}
    \caption{The two EMA paths and their composition at 4 (top) and 64 (bottom) accumulated frames under slow lateral camera motion; the zoomed region is marked on the reference. The accumulated path is noisy at low frame counts but improves with history, while the denoised path is stable but blurry; the variance gate blends the two adaptively (white: prefer denoised, black: prefer accumulated). At 4 frames the not-yet-calibrated gate dips the composite marginally below the denoised path; the dip vanishes within a few frames. PSNR is computed on the zoomed region.}
    \label{fig:dualpath}
    \vspace{-0.6cm}
\end{figure}
Temporal accumulation of raw stochastic samples converges to the alpha-blended reference, but the first few frames are noisy, and suppressing that noise spatially would require aggressive blurring, which lowers the asymptotic quality ceiling once history is long.
We therefore maintain two exponential-moving-average (EMA) chains in parallel: an \textit{accumulated path} over raw samples, which dominates stable regions once history is sufficient, and a \textit{denoised path} over spatially filtered frames, which covers short histories and fills disoccluded regions where reprojection is unreliable. A variance gate blends the two per pixel (Figure~\ref{fig:dualpath}).

\textbf{EMA path update ($\mathcal{U}$).}
At each frame, the previous temporal state is first warped into the current camera view (Sec.~\ref{sec:impl:reproject}). Each path maintains a per-pixel running mean together with an effective sample count, its \textit{history length}. Let $X_{\text{prev}}$ be a path's reprojected mean, $n$ its reprojected history length ($n_s$, $n_h$ for the two paths), $x$ the current frame's value on that path (the spatially filtered colour or the raw stochastic colour) and $t \in [0,1]$ the predicted trust. The state updates as
\begin{equation}
d = n \cdot t + 1, 
X = \frac{X_{\text{prev}} \cdot n \cdot t + x}{d},
n \leftarrow \min(d,\; n_{\max}).
\label{eq:ema-update}
\end{equation}
The trust $t$ interpolates between two extremes: $t = 1$ appends the current frame as one full sample, an ordinary running average; $t = 0$ discards the history and restarts from the current frame; fractional values discount stale history without resetting it, damping ghosting. The denoised path uses a smaller history cap (24 vs.\ 128) for reactivity. A parallel running average of squared luminance on the accumulated path tracks the variance driving the composition gate.

\textbf{Variance-gated composition ($\mathcal{G}_\gamma$) and STAB.}
\label{sec:method:composition}
The final pixel mixes the two paths according to the accumulated path's own uncertainty. Along the accumulated path we maintain a running average of squared luminance $\overline{y^{2}}$ alongside the colour mean $\bar{y}$, giving the temporal variance $\mathrm{var} = \max(\overline{y^{2}} - \bar{y}^{2}, 0)$ and an estimate of the standard error of the mean, $\mathrm{err} = \sqrt{\mathrm{var} / (n_s + \epsilon)}$. The gate weight of the denoised path is
\begin{equation}
g = 1 - r \cdot \bigl(1 - \mathrm{clamp}(\gamma \cdot \mathrm{err},\, 0,\, 1)\bigr),
r = 1 - e^{-n_s / \tau},
\label{eq:gate}
\end{equation}
with a single learned scalar $\gamma$ and a fixed time constant $\tau = 8$ frames; the final colour is $g \cdot D + (1 - g) \cdot S$. Early on, $r \approx 0$ and the denoised path dominates; as history accumulates and the standard error shrinks, the sharp accumulated path takes over. Error spikes such as motion and disocclusion can raise $g$ again, pulling output to the denoised path.

A stabilization pass (\emph{STAB}) suppresses residual temporal jitter of the composited output: a lightweight TAA variant with neighborhood clamping, a uniform blend weight, and a color-consistency rescue against spurious disocclusion resets from depth jitter (details in the appendix).

\textbf{Trust prediction ($f_\theta$).}
Most of the inference budget goes to the accumulated path's trust $t_s$, predicted by a small convolutional network from motion, depth, and photometric-consistency cues; the denoised path's $t_h$ comes from a tiny head biased to distrust history at initialization (architectures in Sec.~\ref{sec:impl:trust}). When the camera is static, we override $t_s$ to 1, where samples are i.i.d.\ and the running mean is the minimum-variance estimator.

\textbf{Spatial filtering ($F_\phi$).}
The spatial filter must be nearly free yet cover a wide footprint: a fixed four-level anisotropic mipmap filter in the spirit of EWA filtering~\citep{zwicker2001ewa}, driven per pixel by learned neural-view scalars $\mathbf{V}=(V_0,V_1,V_2,V_3)$ baked into the surfels (Sec.~\ref{sec:impl:spatial}). The payload is scene-specific and adds 10 floats per Gaussian.

\section{Technical Details}\label{sec:impl}

\subsection{Reprojection and Trust Features}\label{sec:impl:reproject}

The denoiser begins by warping the previous frame's temporal state into the current view: each current pixel is unprojected to world space, reprojected into the previous image plane, and resampled with a 16-tap Catmull-Rom filter, retrieving the previous denoised and accumulated means, the two history lengths, and the two trust maps. Holes left by stochastic sampling (invalid current depth) borrow the average valid depth in a $3\times3$ neighbourhood, falling back to the far plane if the neighbourhood is empty.

From the reprojection we derive four families of \emph{trust features}: \textit{motion} (screen-space motion magnitude, soft-compressed as $m/(m+20)$, and its square), \textit{depth} (relative depth difference and its motion-gated product), \textit{photometric} (luminance and YCoCg distances to the two reprojected means), and \textit{state} (reprojected trust maps, history lengths, and the forward-facing neural-view scalar $V_2$, Sec.~\ref{sec:impl:spatial}). Together with the current observation these form $c_t$ of Eq.~\eqref{eq:unified}; the exact channels are listed in the appendix.

\subsection{History Trust Predictors}\label{sec:impl:trust}

\textbf{Accumulated-path trust $t_s$.}
The predictor is a deep separable convolutional stack with a hidden width of 8: a pointwise $1\times1$ encoder, two depthwise $3\times3$ + pointwise $1\times1$ blocks with SiLU activations, and a pointwise head whose first output channel, passed through a sigmoid, yields $t_s$. Its $5\times5$ receptive field captures local motion boundaries cheaply, and the head bias is initialised to zero ($t_s = 0.5$ at training start).

\textbf{Denoised-path trust $t_h$.}
A tiny head computes $t_h$: a single depthwise $3\times3$ convolution with SiLU followed by a pointwise projection to a scalar and a sigmoid. The output bias is initialised to $-4$ ($t_h \approx 0.02$ at cold start), so the denoised path initially trusts the current frame almost entirely.

The gate scalar $\gamma$ is also learnable.

\subsection{Spatial Anisotropic Mipmap Filtering}\label{sec:impl:spatial}

The filter operates on the 4-channel neural-view tensor $\mathbf{V}$ carried by the surfels, which is the baked form of the per-Gaussian payload $\phi$ of Eq.~\eqref{eq:unified}: $V_0,V_1,V_2$ are the absolute dot products of three learnable per-surfel view vectors with the camera basis (up, right, and forward), $V_0,V_1$ normalized by per-pixel depth and $V_2$ controlling the film-space x-y correlation of the kernel; $V_3$ is a learned scalar controlling blur strength. Trained jointly with the denoiser, these optimal filter footprints are baked into the Gaussian parameters.

A fixed $3\times3$ Gaussian blur of the neural-view tensor yields smoothed scalars from which a $2\times2$ covariance matrix is built; its eigen-decomposition parameterises the filter over a 4-level mipmap pyramid of the current frame's YCoCg colour, probed along the covariance's major axis (exact construction in the appendix). The per-pixel filtering cost is thus bounded by 32 texel reads.

\subsection{Training-Time Details}\label{sec:impl:training}

Training splits into two phases whose costs we account separately: preparing the minimal host model (Stages 1--3), and the joint optimisation of the denoiser and the payload (Stage 4), which is the cost of our contribution. All other training configurations are deferred to the appendix.

\textbf{Minimal host model (Stages 1--3).}
A vanilla 2DGS scene is optimised under the conventional per-Gaussian ordering (Stage 1, 30k steps), cleaned of low-opacity floaters by an opacity-only finetune (Stage 2, 1k steps), and converted to the per-pixel depth-ordered model of Sec.~\ref{sec:method:model} by full finetuning through a depth-peeling differentiable rasterizer (Stage 3, 3k steps). This phase merely converts an off-the-shelf 2DGS reconstruction into our minimal host; a natively view-consistent splatting model would skip it.

\textbf{Joint optimisation (Stage 4).}
All learned components of Eq.~\eqref{eq:unified} are optimised jointly under a single photometric objective (L1 with small SSIM/LPIPS and auxiliary terms, appendix) on the final output $\hat{I}_t$, with the host Gaussian geometry and colour frozen, using the deterministic tile-rendered image as a clean target. Gradients flow through the stochastic renderer, the spatial filter, both trust predictors, and the EMA-gate composition in a single graph; optimisation proceeds with history states detached for training efficiency. Temporal behaviour is learned on synthetic camera chains whose length follows a curriculum up to 128 frames, matching the inference-time history cap (chain synthesis and trust-shaping regularisers in the appendix). The payload $\phi$ is baked per Gaussian and therefore trains per scene; the shared $\theta$ adapts to a new scene with a short finetune (15k steps on the source scene, 4k steps on transferred scenes, ${\sim}1$ hour), and can be deployed with no per-scene denoiser training at some quality cost (zero-shot transfer, Sec.~\ref{sec:results:zeroshot}).

\section{Experiments and Ablations}
\label{sec:experiments}

\subsection{Implementation Details}
We train with PyTorch and SlangPy~\citep{slangpy} and implement the integrated renderer with NVRHI~\citep{nvrhi} (Vulkan backend); all experiments run on an NVIDIA RTX~3090 GPU. All scenes share the same hyper-parameters (details in Appendix) and the two-phase training recipe of Sec.~\ref{sec:impl:training} (room as the source scene). We provide code and running instructions for reproduction in the supplements.

\subsection{Experimental Setup}
\textbf{Datasets.} We evaluate on the main release (7 scenes) of MipNeRF360~\citep{barron2022mipnerf360} and the training-data subset of TanksAndTemples~\citep{tanksandtemples}. 
All renderers are evaluated at the native training resolution (resized to $1600$\,px image width on both datasets); we cap the Gaussian count at 3-million for training efficiency. 

\textbf{Baselines.} We compare against the deterministic vanilla 2DGS tile renderer~\citep{huang20242dgs}, which also renders the reference images for all metrics, and against an adapted implementation of stochastic rasterization pipeline from StochasticSplats~\citep{kheradmand2025stochasticsplats}. StochasticSplats originally renders a 3DGS scene finetuned under StopThePop rasterization~\citep{radl2024stopthepop} which differs from our minimal 2DGS models. 
We render our per-pixel 2DGS scenes with the same stochastic rasterizer that feeds our denoiser, and reimplement the temporal anti-aliasing (TAA) accumulation described in that work on its output. We refer to this adapted baseline as \emph{ST-TAA} below: baseline and denoiser consume the identical 1\,spp stream of the identical scene representation, so any quality difference is attributable to the denoiser alone. We further compare against SVGF~\citep{schied2017svgf}, classic real-time Monte Carlo denoiser, minimally adapted to the stochastic stream: 3x3 depth hole-fix, temporal accumulation with per-pixel luminance moments, followed by an \`a-trous wavelet filter edge-stopped by depth and variance-guided luminance, with hand-tuned parameters (details in Appendix). On static cameras, ST-TAA reduces to equal-weight frame accumulation; we additionally report a 1024-sample reference accumulation (\emph{Ref.\ (1024)}), Monte-Carlo mean of 1024 independent 1\,spp renders of the same view (or of each frame's exact pose along navigation tracks), which upper-bounds what any temporal filter can extract from the stochastic stream.

\textbf{Metrics.} We report PSNR, SSIM, and LPIPS(VGG), plus ColorVideoVDP~\citep{mantiuk2024colorvideo} (JOD units) (CVVDP) for navigation video quality (synthesized from every 10th track frame of the $\sim$300 fps recordings for evaluation speed, $\sim$30 fps). Temporal metrics are averaged over every 10th frame of each navigation track, using the vanilla tile-rendered images as the reference.

\subsection{Static Quality on Benchmark Datasets}
\begin{table}[t]
\centering
\footnotesize
\setlength{\tabcolsep}{2pt}
\begin{tabular}{@{}l ccc ccc@{}}
\toprule
 & \multicolumn{3}{c}{MipNeRF360} & \multicolumn{3}{c}{TanksAndTemples} \\
\cmidrule(lr){2-4} \cmidrule(lr){5-7}
Method & PSNR$\uparrow$ & SSIM$\uparrow$ & LPIPS$\downarrow$ & PSNR$\uparrow$ & SSIM$\uparrow$ & LPIPS$\downarrow$ \\
\midrule
2DGS & 29.45 & 0.875 & 0.199 & 21.50 & 0.700 & 0.379 \\
Ref.\ (1024) & 28.25 & 0.851 & 0.202 & 20.69 & 0.675 & 0.388 \\
\hline \vspace{-0.22cm}\\
ST-TAA (128) & 27.92 & \textbf{0.822} & \textbf{0.219} & \textbf{20.61} & \textbf{0.635} & \textbf{0.431} \\
Ours (16) & 26.89 & 0.741 & 0.330 & 20.33 & 0.548 & 0.496 \\
Ours (98) & \textbf{27.94} & \textbf{0.822} & 0.226 & 20.60 & 0.631 & 0.436 \\
\midrule
Ours (128) & 28.02 & 0.829 & 0.217 & 20.63 & 0.640 & 0.428 \\
Ours (z.s. 128) & 28.03 & 0.829 & 0.217 & 20.62 & 0.639 & 0.428 \\
\bottomrule
\end{tabular}
\caption{Static quality, averaged per dataset; per-scene PSNR is reported in the appendix. \emph{Vanilla 2DGS} is the deterministic tile renderer~\citep{huang20242dgs}. \emph{Ref.\ (1024)} is the Monte-Carlo accumulation reference (Sec.~\ref{sec:experiments}). \emph{ST-TAA} is the adapted StochasticSplats baseline (Sec.~\ref{sec:experiments}), accumulated over 128 static frames. \emph{Ours} renders through the learned denoiser at 16, 98, or 128 accumulated frames (spp counts in parentheses); the 98\,spp point approximately matches ST-TAA (128) in wall-clock time, making it comparable (Table~\ref{tab:performance}); bold marks the better of this equal-time pair in each column (both when equal after rounding). \emph{Ours (z.s.)} deploys the room checkpoint on every scene with no per-scene denoiser training (zero-shot transfer) (Sec.~\ref{sec:results:zeroshot}).%
}
\label{tab:static_main}
\vspace{-0.5cm}
\end{table}

Table~\ref{tab:static_main} reports static quality on the two benchmarks (per-scene numbers in the appendix). 
The remaining gap from the 1024 sample Monte-Carlo mean (Ref.) to the deterministic renderer is the residual cost of converting the geometry to the per-pixel model (Sec.~\ref{sec:method:model}).
On a static camera any temporal filter can only converge to the reference. Our 128-frame output wins a small margin at equal samples. The win comes from the variance gate, which admixes a small fraction of the spatially filtered denoised path into the accumulated color. 
From a practical operating point, what matters is how quickly the output converges. Our 16-frame output closes most of the gap to the reference, coming within 1.1\,dB of its 128-frame quality, and converges markedly faster than ST-TAA because its denoised path gathers spatially available samples (convergence curves in the appendix). On TanksAndTemples, where all methods are trained on the full capture of each scene, the 128-frame PSNR likewise closes to the bound (20.63 vs.\ 20.69\,dB), with a small residual SSIM/LPIPS gap.

\subsection{Temporal Quality under Free Navigation}
\begin{table}[t]
\centering
\footnotesize
\setlength{\tabcolsep}{3.5pt}
\begin{tabular}{@{}l cccc|c@{}}
\toprule
Method & PSNR$\uparrow$ & SSIM$\uparrow$ & LPIPS$\downarrow$ & CVVDP$\uparrow$ & ms$\downarrow$ \\
\midrule
Ref.\ (1024) & 32.96 & 0.950 & 0.080 & 8.73 & -- \\
\midrule
Raw 1\,spp & 18.26 & 0.282 & 0.671 & 4.94 & 2.71 \\
SS 1\,spp$^{\dagger}$ & 17.86 & 0.292 & 0.659 & 4.67 & 3.2$^{\dagger}$ \\
SS 4\,spp$^{\dagger}$ & 22.54 & 0.526 & 0.540 & 6.87 & 6.5$^{\dagger}$ \\
ST-TAA & 21.48 & 0.418 & 0.615 & 5.95 & 2.88 \\
SVGF & 25.57 & 0.785 & 0.297 & 6.29 & 2.86 \\
Ours & \textbf{29.80} & 0.867 & 0.244 & \textbf{7.57} & 3.75 \\
Ours (z.s.) & 29.34 & \textbf{0.873} & \textbf{0.223} & 7.38 & 3.64 \\
\bottomrule
\vspace{-0.5cm}
\end{tabular}
\caption{Quality under free camera navigation on MipNeRF360, averaged over all scene tracks (per-track numbers in the appendix), using the 2DGS tile renderer as the reference. Raw 1\,spp is the single-sample stochastic output; ST-TAA is the adapted StochasticSplats baseline, temporal accumulation without learned trust or spatial filtering; SVGF is the adapted variance-guided filtering baseline (see appendix). Ref.\ (1024) is the per-frame Monte-Carlo ceiling, not a practical method. The last column is the mean per-frame time over the seven tracks on an RTX~3090. $^{\dagger}$SS 1\,spp/4\,spp are the official StochasticSplats model run on its own finetuned 3DGS scenes over the same camera tracks for 1 or 4 spp, measured against its own 1024\,spp accumulation as the reference (no ground truth exists on free tracks); it is listed to show that the 1\,spp noise level is inherent with the stochastic formulation. Its per-frame error distribution closely tracks our Raw 1\,spp row (per-frame PSNR correlation $r=0.90$, and its CVVDP score likewise matches our Raw 1\,spp row). SS timings come from the official viewer. \emph{Ours (z.s.)} deploys the room checkpoint on every scene with no per-scene denoiser training (Sec.~\ref{sec:results:zeroshot}).}
\label{tab:navigation}

\vspace{-0.4cm}
\end{table}

\begin{figure}[t]
\centering
\setlength{\tabcolsep}{1.5pt}
\scriptsize
\begin{tabular}{@{}cccc@{}}
\includegraphics[width=0.30\columnwidth]{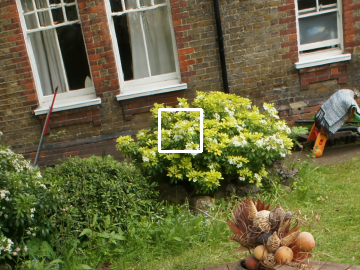} &
\includegraphics[width=0.225\columnwidth]{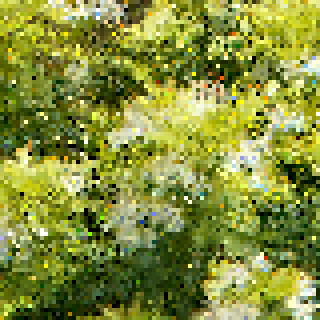} &
\includegraphics[width=0.225\columnwidth]{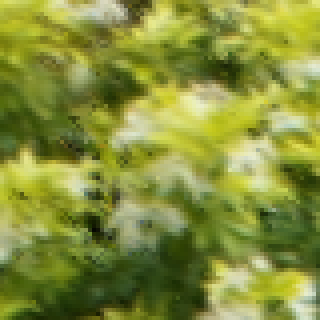} &
\includegraphics[width=0.225\columnwidth]{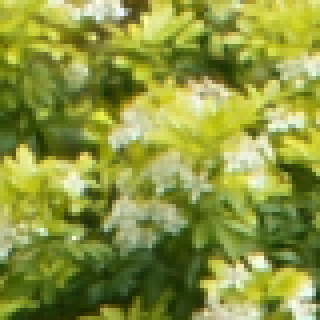} \\
\includegraphics[width=0.30\columnwidth]{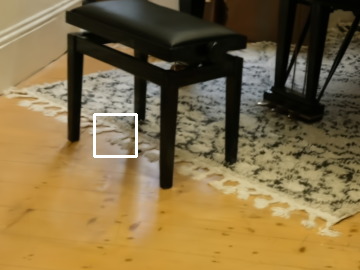} &
\includegraphics[width=0.225\columnwidth]{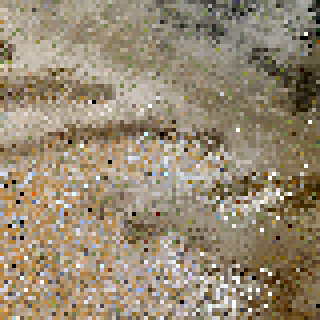} &
\includegraphics[width=0.225\columnwidth]{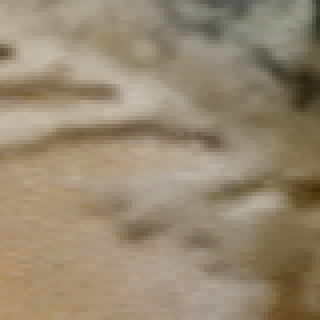} &
\includegraphics[width=0.225\columnwidth]{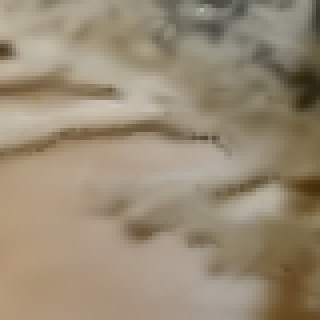} \\
2DGS (frame) & ST-TAA & Ours & 2DGS (zoom) \\
\vspace{-0.4cm}
\end{tabular}
\caption{Garden (top) and Room (bottom) navigation tracks. Left: vanilla-rendered reference crop with the zoom region marked; right: the region for each renderer. 
}
\label{fig:navigation_qual}
\end{figure}

Free navigation is a common real-time workload. We record camera tracks by hand in the interactive viewer on all seven MipNeRF360 scenes (provided in our supplements), mimicking how a user actually navigates, and compare five renderers: raw 1\,spp stochastic output, ST-TAA, SVGF, our denoiser, and the vanilla tile renderer as the reference. Table~\ref{tab:navigation} averages the metrics over the seven tracks (per-track numbers in the appendix). Our denoiser stays close to the deterministic reference, while ST-TAA fails to suppress the Monte Carlo noise and leaves heavy, clearly visible flickering under camera motion (Figure~\ref{fig:navigation_qual}). Its collapse is not a convergence limit but a failure to retain valuable temporal history, which is what our learned trust prediction addresses. SVGF fares markedly better than plain TAA, but without noise-free guidance features its fixed edge-stopping strategy is severely interfered by floating Gaussians and holes in scenes. The Ref.\ (1024) ceiling sits well above every temporal method under navigation, whereas on static cameras our denoiser essentially reaches it (Table~\ref{tab:static_main}); the residual gap thus comes from reusing history across poses: reprojection mismatch and disocclusions.

\subsection{Zero-Shot Transfer to Unseen Scenes}
\label{sec:results:zeroshot}
The trust predictors consume mostly relative cues of the stochastic error process, so the learned trust partially transfers across scenes. We test this by copying the room checkpoint verbatim to the other scenes, neutralising the scene-specific neural-view payload: the trust feature $V_2$ fed to the trust estimator is set to zero, and the spatial filter falls back to a fixed isotropic kernel ($a_x = a_y = 2$, $\rho = 0$).
In tables~\ref{tab:static_main} and~\ref{tab:navigation}, static quality of \emph{Ours (z.s.)} matches the fine-tuned model, while under navigation the zero-shot variant gives up some PSNR/CVVDP margin yet slightly improves SSIM and LPIPS: the fixed kernel filters more aggressively than the learned kernels, which is favored by the perceptual metrics. Either way the differences are small next to the margin over ST-TAA. Per-track numbers are in the appendix.

\subsection{Performance Analysis}
\begin{table}[t]
\centering
\small
\begin{tabular}{l c}
\toprule
Method (ms/frame) & 7-track mean $\downarrow$ \\
\midrule
2DGS ~\cite{huang20242dgs} & 17.96 \\
2DGS (hardware raster.) & 7.99 \\
Raw 1\,spp & 2.71 \\
\quad + TAA (ST-TAA) & 2.88 \\
Ours (full pipeline) & 3.75 \\
\bottomrule
\end{tabular}
\caption{Per-frame rendering time, averaged over the seven navigation tracks (per-scene numbers in the appendix). The 2DGS row is the official implementation measured with auxiliary channel outputs removed; the hardware raster. row is our hardware-rasterized reimplementation of 2DGS, included as a stronger baseline. The cost of denoiser is nearly constant across scenes.}
\label{tab:performance}
\vspace{-0.6cm}
\end{table}

Table~\ref{tab:performance} reports per-frame rendering time averaged over the seven navigation tracks (per-scene numbers in the appendix). Raw stochastic rasterization is $2.9\times$ faster than the sorted alpha-blending renderer on the same hardware-rasterization pipeline, and the full denoising pipeline keeps a $2.1\times$ advantage: the denoiser adds ${\sim}1$\,ms per frame, nearly constant across scenes. The speedup varies considerably with scene content, from $1.2\times$ (kitchen, where blending overdraw is low) to $3.5\times$ (bicycle), consistent with the scaling behaviour of sorted blending (Sec.~\ref{sec:method:overview}), and stays well above the denoiser's overhead.

\subsection{Ablations}
\label{sec:ablations}
\begin{table}[t]
\centering
\small
\begin{tabular}{l ccc}
\toprule
Variant & PSNR$\uparrow$ & SSIM$\uparrow$ & LPIPS$\downarrow$ \\
\midrule
Full model                        & \textbf{26.63} & \textbf{0.794} & \textbf{0.238} \\
\midrule
w/o $t_s$ prediction ($t_s{\equiv}1$) & 22.51 & 0.695 & 0.323 \\
w/o $t_s$ prediction ($t_s=\mathrm{heur.}$)  & 25.82 & 0.755 & 0.268 \\
w/o spatial filter                & 23.52 & 0.676 & 0.359 \\
w/o stabilization pass            & 25.94 & 0.736 & 0.311 \\
\bottomrule
\end{tabular}
\caption{Ablations on the garden navigation track (evaluated on every 10th frame of the original 1563 frames, metrics against the vanilla tile renderer). The learned trust prediction is the most impactful single component, followed by the spatial filter. The stabilization pass improves all three metrics and, more importantly, removes most of the visible temporal jitter at the cost of slight ghosting. For the $t_s=\mathrm{heur.}$ ablation, we replace the neural network with $t_s = \exp(-\sigma \cdot \frac{|d_1 - d_2|}{\max(d_1, d_2)})$ heuristic as a stronger baseline, where $d_1, d_2$ are pixel depths of the reprojected/current pixels and $\sigma$ is grid-searched on the same track for best performance (oracle). }
\label{tab:ablation_studies}
\vspace{-0.6cm}
\end{table}

Table~\ref{tab:ablation_studies} ablates the learned trust prediction, the spatial filter, and the final stabilization pass on the garden navigation track. The learned trust is the largest single component: forcing $t_s \equiv 1$ approximates a naive 'always accumulate' strategy. The spatial filter is the second contributor, providing filtered results in disoccluded or recently revealed regions. The stabilization pass improves all three per-frame metrics.

\section{Discussion, Limitations, and Conclusion}
\textit{Balancing Temporal Stability And Ghosting.}
We found that the final output can still exhibit visible flicker. We have to use STAB to trade temporal stability with some minor ghosting. 
We observe lower-than-expected predicted trust within flickering regions, which possibly relates to the limited capacity of predictor networks. 
To alleviate this issue, it is possible to further enhance the denoising path, 
or reduce the variance of the stochastic output distribution itself. 

\textit{Performance across different rendering paths.}
Our current realization of the whole pipeline does not match the absolute single-channel throughput of highly optimised sorted-blending engines~\citep{feng2025flashgs}, order-independent compositors~\citep{du2026mobilegs}, or compact surfel-hybrid representations~\citep{ye2025ges}, and we make no claim of superiority over these differently engineered pipelines. 

However, the comparison also shifts with scene characteristics and host implementation. Sorted and blended pipelines pay per-fragment costs that grow with primitive count, overdraw, and the number of output channels, whereas the stochastic pipeline pays a roughly per-pixel cost plus a fixed denoiser overhead; recent stochastic renderers already drive hundreds of millions of Gaussians interactively~\citep{Rijsdijk2026GaussianPointSplatting}, where sorted pipelines struggle. Within this landscape, our contribution is deliberately narrow: removing the Monte-Carlo noise that has so far blocked the stochastic route under free navigation, at a ${\sim}1$ms per-frame overhead. 

\textit{Composability and future work.}
These directions compose rather than compete. Because the denoiser's input contract is only the view-consistent stochastic visibility stream of Sec.~\ref{sec:method:stream}, it can in principle sit behind a culled or more compact representation~\citep{ye2025ges} or a faster implementation of stochastic rasterization \citep{Rijsdijk2026GaussianPointSplatting}. On the engineering side, the stochastic rasterization natively produces a noisy depth buffer at zero extra cost, which can potentially enable hierarchical Z-culling~\citep{greene1993hierarchical}. Integration is left to future work.

In all, we present a lightweight temporal neural denoiser that makes view-consistent stochastic Gaussian splatting practical for free-navigation rendering while preserving the efficiency of sort-free, blend-free rasterization. By decoupling the stream denoising problem from a specific splatting representation, our approach offers a practical postprocessing stage for future stochastic renderers.

\bibliographystyle{plainnat}
\bibliography{references}

\clearpage
\appendix
\clearpage

This appendix provides additional implementation details, extended quantitative and qualitative evaluations, and further experiments that complement the main text.

\section{Predictor Input Channels}\label{app:inputs}

Tables~\ref{tab:stepa-inputs} and~\ref{tab:stepc-inputs} list the exact per-channel inputs of the two trust predictors. All channels are per-pixel maps assembled during reprojection. Please refer to the code for more details.

\paragraph{Accumulated-path predictor architecture.}
The full layer sequence is: pointwise $1{\times}1$ encoder $\to$ depthwise $3{\times}3$ + pointwise $1{\times}1$ $\to$ pointwise $1{\times}1$ $\to$ depthwise $3{\times}3$ + pointwise $1{\times}1$ $\to$ pointwise head, all with SiLU activations and hidden width 8. The main text's description of ``two depthwise $3{\times}3$ + pointwise $1{\times}1$ blocks'' elides the intermediate pointwise layer between the two depthwise blocks for brevity.

\begin{table}[h]
\centering
\small
\begin{tabular}{cl}
\toprule
Ch. & Input \\
\midrule
0 & $|Y_{\text{cur}} - Y_{\text{stoch}}|$: luma diff to reprojected accumulated mean \\
1 & reprojected previous $t_s$ \\
2 & squared compressed motion magnitude \\
3 & accumulated-path history length (normalised) \\
4 & compressed motion magnitude \\
5 & relative depth difference \\
6 & motion-gated relative depth difference \\
7 & forward-facing neural-view scalar ($V_2$) \\
\bottomrule
\end{tabular}
\caption{Accumulated-path trust predictor ($t_s$) inputs, 8 channels.}
\label{tab:stepa-inputs}
\end{table}

\begin{table}[h]
\centering
\small
\begin{tabular}{cl}
\toprule
Ch. & Input \\
\midrule
0 & $|Y_{\text{cur}} - Y_{\text{den}}|$: luma difference to reprojected denoised mean \\
1 & reprojected denoised-path luma \\
2 & reprojected previous $t_h$ \\
3 & denoised-path history length (normalised) \\
4 & compressed motion magnitude \\
5 & YCoCg distance to reprojected denoised mean \\
6 & relative depth difference \\
7 & squared compressed motion magnitude \\
\bottomrule
\end{tabular}
\caption{Denoised-path trust head ($t_h$) inputs, 8 channels.}
\label{tab:stepc-inputs}
\end{table}

\section{Spatial Filter Details}\label{app:filter}

Exact construction of the covariance matrix in the spatial-filtering section of the main text. A fixed $3\times3$ Gaussian blur of the 4-channel neural-view tensor produces smoothed scalars $(V'_0, V'_1, V'_2, V'_3)$. Per-axis scales $a_x = \mathrm{relu}(V'_1) + \mathrm{softplus}(V'_3)$ and $a_y = \mathrm{relu}(V'_0) + \mathrm{softplus}(V'_3)$ (the softplus floor guarantees a minimum blur) and a normalised correlation $\rho = \tanh(V'_2)$ form the covariance matrix $\bigl[\begin{smallmatrix} a_x & z \\ z & a_y \end{smallmatrix}\bigr]$ with $z = \rho\sqrt{a_x a_y + \epsilon}$. Its eigen-decomposition gives major/minor axis lengths $\lambda_1, \lambda_2$ and orientation $\theta$. The minor axis selects the mip level $\mathrm{clamp}(\log_2 \lambda_2,\, 0,\, 3)$; four probes at offsets $\{-1.5, -0.5, 0.5, 1.5\}\times\lambda_2$ along the major axis are trilinearly sampled and averaged with weights $\omega_i = \exp\!\bigl(-\tfrac{1}{2}(o_i \lambda_2 / \lambda_1)^2\bigr)$.

\section{Stabilization Pass (STAB) Details}\label{app:stab}

\textbf{Purpose.}
The trust for the two EMA chains is not sharp enough, so large regions in the composited output retain a characteristic residual noise: temporally high-frequency (per-frame flicker that does not converge), spatially low-frequency ($\sim$3\,px blobs), and small in amplitude ($\sim$0.05 in $[0,1]$) but still visually annoying. Derived from TAA, STAB is a post-pass on the composited image that removes this jitter. It is bypassed whenever the camera is static (tolerance-based detection on the camera matrix).

\textbf{Design.}
Let $C$ be the current composited colour and $H$ the previous stabilized frame, reprojected into the current view with Catmull--Rom sampling (which the reprojection stage already uses for the EMA chains). STAB computes
\begin{equation}
\mathrm{out} = a \cdot \mathrm{clamp}(H,\; C - \mathrm{tol},\; C + \mathrm{tol}) + (1 - a) \cdot C,
\label{eq:stab}
\end{equation}
where the neighbourhood tolerance $\mathrm{tol} = \tfrac{1}{2}(C_{\max}^{3\times3} - C_{\min}^{3\times3}) + s$ is computed per channel from the current frame's $3\times3$ neighbourhood (already available from history rectification) with slack $s = 0.04$; the clamp bounds the ghosting amplitude to the slack level. The blend weight is a fixed $a = 0.9$, except that a depth-disocclusion test re-seeds the history ($a = 0$) when the relative depth change exceeds $0.1$, ensuring recovery on genuine surface changes.

\textbf{Rescue (Color fix).}
Because the stochastic depth winner jitters even on stable surfaces, the depth re-seed false-fires on many pixels that could safely be smoothed. A rescue signal may override the re-seed (OR-logic): the \emph{Color fix} keeps blending when the clamped history is already close to the current frame, $\lVert C - \mathrm{clamp}(H) \rVert_\infty < 0.10$, a direct visual-difference test that is unlikely to fire on real motion edges; it can cause minor ghosting. Since the clamp caps ghosting regardless, rescuing is mostly safe. The output of Eq.~\eqref{eq:stab} becomes both the displayed frame and the next frame's history. The cost of STAB is below measurement noise.

\section{Training Configuration}\label{app:hyper}

\textbf{Two-phase recipe.}
Table~\ref{tab:stages} summarises the training pipeline (main text, Training-Time Details). Stages 1--3 prepare the minimal host model: Stages 1--2 optimise a vanilla 2DGS scene under the conventional per-Gaussian ordering, and Stage 3 converts it to the per-pixel depth-ordered model of the main text by full finetuning through a depth-peeling differentiable rasterizer. Stage 4 trains the denoiser jointly with the neural-view payload under the stochastic renderer, with the remaining Gaussian parameters frozen.

\begin{table}[h]
\centering
\small
\begin{tabular}{cllc}
\toprule
Stage & Renderer & Optimised & Steps \\
\midrule
1 & Tile Rasterizer & all & 30k \\
2 & Tile Rasterizer & opacity & 1k \\
3 & Depth Peeler & full finetune & 3k \\
4 & Stochastic Rasterizer & denoiser + $V_{0,1,2,3}$ & 15k / 4k \\
\bottomrule
\end{tabular}
\caption{Two-phase training recipe (four stages). Stage 4 trains 15k steps on room and transfers to every other scene with 4k steps, starting from the room denoiser. The 4k transfer is our default per-scene refinement. The room checkpoint can also be deployed with no per-scene denoiser training at some quality cost (main text, Zero-Shot Transfer).}
\label{tab:stages}
\end{table}

\textbf{Host model stages.}
Stage 1 trains vanilla 2DGS for 30k steps without normal regularisation ($\lambda_n = 0$), on par with the original 2DGS results; enabling the normal regulariser costs about 1\,dB because it constrains surfel orientation. Stage 2 finetunes opacity only for 1k steps (penalty weight $0.003$, prune threshold $0.05$) to remove low-opacity floaters without touching geometry.

\textbf{Objective.}
The final-frame loss combines L1, SSIM ($0.05$) and LPIPS ($0.01$) against the tile-rendered vanilla image—a clean, deterministic target—plus an auxiliary L1 term (weight $0.2$) pulling the accumulated path toward the vanilla colour.

\textbf{Synthetic temporal chains.}
Since no video sequences exist for the training views, temporal behaviour is learned on synthetic camera chains: 70\% of steps use temporal history, chain lengths follow a curriculum from 2 to 128 frames, and only the last 8 frames of each chain are rendered stochastically while older frames come from the cheap tile renderer (\textit{fake start}), so a 128-frame chain costs about nine stochastic renders. Camera perturbations are sampled from five classes—lateral dolly, drift, large, small, and near-static ($1/15$ scale)—covering both smooth motion and disocclusion.

\textbf{Trust shaping.}
Three small mechanisms stabilise the learned trust: (i) near-static regularisers pull $t_s \!\to\! 1$ and $t_h \!\to\! 1$ (both weights $0.01$; fast transfer uses $0.005$ for both and adds a disocclusion penalty of weight $0.05$ on both) on near-static chains, encouraging full accumulation when nothing moves; (ii) an authoritative disocclusion mask computed from the tile renderer's depth and opacity hard-zeroes predicted trust in genuinely disoccluded pixels during state updates, with an L1 penalty (weight $0.03$) suppressing trust there during training; and (iii) the composition gate is forced to the denoised path for the first 2000 steps, so the gate scalar $\gamma$ is calibrated only after both paths produce meaningful outputs. A symmetry regulariser (weight $0.01$) keeps the spatial filter's axis scales consistent.

\section{SVGF Baseline Details}\label{app:svgf}

Our SVGF baseline follows Schied et al.'s SVGF: temporal accumulation with first/second luminance moments, then \`a-trous wavelet filtering edge-stopped by the estimated temporal variance and geometry. It runs in the same Vulkan pipeline as our denoiser and consumes the identical 1\,spp stream. The stochastic stream provides no noise-free G-buffer, so the following minimum adaptations are made:

\begin{itemize}
\item \textbf{Hole-fixed depth.} Miss pixels of the 1\,spp visibility pass carry no valid depth (flagged by zero accumulated alpha); they are filled from the nearest valid $3{\times}3$ neighbour before any depth use, and pixels with no valid neighbour fall back to the far plane. All reprojection and edge-stopping read this hole-fixed depth.
\item \textbf{Temporal accumulation.} Per pixel, the previous frame's integrated colour and moments are reprojected (current depth $\to$ world $\to$ previous camera, Catmull-Rom history sampling) and blended exponentially: $c \leftarrow (1-\alpha)\, c_{\mathrm{hist}} + \alpha\, c_{\mathrm{cur}}$ with $\alpha = 0.02$, likewise for the luminance moments $m_1, m_2$. A pixel is treated as disoccluded (reset to the current frame) when fewer than 2 of the 4 nearest previous-depth texels pass the relative-depth test with threshold $0.4$. The feedback history is the pre-filter integrated colour, as in classic SVGF.
\item \textbf{Spatial filter.} A single \`a-trous iteration (stride 1) of the $3{\times}3$ binomial kernel $\frac{1}{16}[1\,2\,1; 2\,4\,2; 1\,2\,1]$ (we tried multiple iterations and it blurs the output violently, causing significant visual quality downgrade. Thus, we only kept 1 iteration.). Per-tap weights combine a relative-depth bilateral term against the $3{\times}3$-smoothed centre depth, $\exp\!\bigl(-|\Delta z| / (\sigma_z \bar{z} + 10^{-2})\bigr)$ with $\sigma_z = 0.3$, and the variance-guided luminance term $\exp\!\bigl(-|\Delta l| / (\sigma_l \sqrt{\mathrm{var}} + 10^{-2})\bigr)$ with $\sigma_l = 8$, where $\mathrm{var} = \max(0, \bar{m}_2 - \bar{m}_1^2)$ is estimated from the $3{\times}3$-averaged moments. The paper's depth-gradient and normal terms are dropped: the 1\,spp winner depth jitters between surfaces, so screen-space depth gradients are noise (they quantise the filter taps into a visible grid), and no surface normals exist.
\end{itemize}

All parameters ($\alpha = 0.02$, one \`a-trous iteration, $\sigma_l = 8$, $\sigma_z = 0.3$, disocclusion threshold $0.4$) were hand-tuned on the interactive viewer for the best visual trade-off between residual noise and blur; per-track numbers are included in Table~\ref{tab:navigation_full}.

\section{Per-Scene Rendering Performance}\label{app:perf}

Table~\ref{tab:perf_scenes} reports the per-scene per-frame times underlying the main text's performance table, measured on the navigation tracks (RTX~3090, median of three pipelined runs). Sorted alpha blending varies widely with scene content ($4.5$--$13.7$\,ms), while the full stochastic pipeline stays in a narrow band ($3.4$--$4.0$\,ms); the denoiser's marginal cost over raw rasterization is about $1.0$\,ms on every scene. Table~\ref{tab:perf_scenes_extra} adds the SVGF baseline and the zero-shot variant, which share the same stochastic rasterization front end and differ only in the post-processing stage.

\begin{table}[h]
\centering
\small
\setlength{\tabcolsep}{3.5pt}
\begin{tabular}{l cc}
\toprule
Scene & SVGF & Ours (zero-shot) \\
\midrule
bicycle & 2.94 & 3.76 \\
bonsai  & 2.63 & 3.44 \\
counter & 2.91 & 3.66 \\
garden  & 2.97 & 3.75 \\
kitchen & 3.00 & 3.76 \\
room    & 3.04 & 3.79 \\
stump   & 2.56 & 3.35 \\
\midrule
mean    & 2.86 & 3.64 \\
\bottomrule
\end{tabular}
\caption{Per-scene per-frame rendering time (ms) for the SVGF baseline and the zero-shot variant, measured under the identical protocol as Table~\ref{tab:perf_scenes}. Both consume the same 1\,spp stochastic stream; SVGF replaces the learned post-processing with hand-tuned variance-guided filtering, and the zero-shot variant skips per-scene denoiser training.}
\label{tab:perf_scenes_extra}
\end{table}

\begin{table}[h]
\centering
\small
\setlength{\tabcolsep}{3.5pt}
\begin{tabular}{l cccccc}
\toprule
Scene & 2DGS & 2DGS (hw) & Raw & ST-TAA & Ours & Speedup \\
\midrule
bicycle & 29.32 & 13.73 & 2.86 & 3.03 & 3.95 & 3.5$\times$ \\
bonsai  & 17.66 &  8.14 & 2.49 & 2.66 & 3.52 & 2.3$\times$ \\
counter & 16.24 &  6.33 & 2.74 & 2.89 & 3.75 & 1.7$\times$ \\
garden  & 16.27 &  6.57 & 2.79 & 2.97 & 3.85 & 1.7$\times$ \\
kitchen & 10.38 &  4.49 & 2.84 & 3.01 & 3.86 & 1.2$\times$ \\
room    & 13.94 &  6.18 & 2.89 & 3.04 & 3.88 & 1.6$\times$ \\
stump   & 21.91 & 10.47 & 2.38 & 2.53 & 3.44 & 3.0$\times$ \\
\midrule
mean    & 17.96 &  7.99 & 2.71 & 2.88 & 3.75 & 2.1$\times$ \\
\bottomrule
\end{tabular}
\caption{Per-scene per-frame rendering time (ms) on the navigation tracks. \emph{2DGS} is the official CUDA tile renderer with auxiliary channels removed; \emph{2DGS (hw)} is our hardware-rasterized reimplementation of the same sorted alpha-blending algorithm; \emph{Raw} is 1\,spp stochastic rasterization; \emph{ST-TAA} adds the baseline temporal accumulation; \emph{Ours} is the full denoising pipeline. Speedup is 2DGS (hw) over Ours.}
\label{tab:perf_scenes}
\end{table}

\section{Static Convergence Curve}\label{app:convergence}

\begin{figure}[t]
\centering
\includegraphics[width=0.7\linewidth]{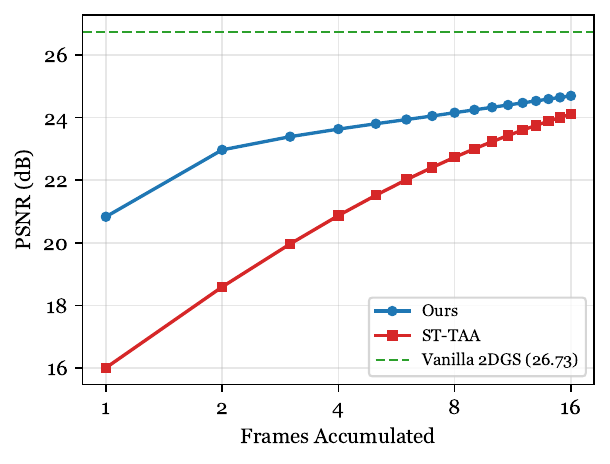}
\caption{Static-view convergence on garden. PSNR versus the number of accumulated frames on a static camera, averaged over the 24 held-out views. The dashed line marks the deterministic vanilla 2DGS renderer, which renders each frame identically and independently of accumulation.}
\label{fig:static_convergence}
\end{figure}

Figure~\ref{fig:static_convergence} shows how static-view quality builds up over the first accumulated frames. Each point renders the same static view for $k$ consecutive frames and reports PSNR against ground truth, averaged over the 24 held-out views of garden. On static views ST-TAA reduces to equal-weight Monte-Carlo accumulation. We verified its curve coincides with an explicit reference accumulator to within $0.02$\,dB over the whole range. Our denoiser converges markedly faster at low frame counts because its spatially filtered denoised path carries the output before temporal history builds up, and then stays on the same Monte-Carlo convergence trajectory as plain accumulation.

\section{Variance Gate Ablation}\label{app:gate_ablation}

Table~\ref{tab:gate_ablation} ablates the variance gate on the garden navigation track: forcing the per-pixel composition weight $g$ to either path exclusively is strictly worse than the learned gate on all three metrics.

\begin{table}[t]
\centering
\small
\begin{tabular}{l ccc}
\toprule
Variant & PSNR$\uparrow$ & SSIM$\uparrow$ & LPIPS$\downarrow$ \\
\midrule
Full model (learned $g$) & \textbf{26.63} & \textbf{0.794} & \textbf{0.238} \\
\midrule
$g \equiv 0$ (accumulated path only) & 25.14 & 0.751 & 0.298 \\
$g \equiv 1$ (denoised path only) & 25.95 & 0.752 & 0.247 \\
\bottomrule
\end{tabular}
\caption{Variance-gate ablation on the garden navigation track (same protocol as the ablation table in the paper). $g$ is the per-pixel weight of the denoised path in the final composition in the main text; forcing it to either path exclusively degrades all three metrics. The learned gate improves over the better fixed extreme by 0.68\,dB PSNR.}
\label{tab:gate_ablation}
\vspace{-0.6cm}
\end{table}

\section{Per-Scene and Per-Track Results}\label{app:static}

Tables~\ref{tab:static_mipnerf} and~\ref{tab:static_tnt} report the per-scene PSNR underlying the main text's static-quality averages, and Table~\ref{tab:navigation_full} reports the per-track numbers underlying the main text's navigation table. Table~\ref{tab:ss_tracks} reports the per-track breakdown of the official StochasticSplats 1\,spp/4\,spp rows. Table~\ref{tab:zeroshot_tracks} reports the per-track breakdown of the main text's zero-shot transfer experiment. The hand-recorded navigation camera tracks used for all track evaluations are provided in the code supplement under \texttt{data/tracks/}.

\begin{table*}[t]
\centering
\small
\begin{tabular}{l ccccccc ccc}
\toprule
 & \multicolumn{7}{c}{PSNR$\uparrow$ per scene} & \multicolumn{3}{c}{Average} \\
\cmidrule(lr){2-8} \cmidrule(lr){9-11}
Method & bicycle & bonsai & counter & garden & kitchen & room & stump & PSNR$\uparrow$ & SSIM$\uparrow$ & LPIPS$\downarrow$ \\
\midrule
2DGS & 24.71 & 32.12 & 28.82 & 26.73 & 32.49 & 31.53 & 29.74 & 29.45 & 0.875 & 0.199 \\
Ref.\ (1024) & 24.20 & 30.65 & 28.14 & 26.15 & 30.24 & 30.51 & 27.88 & 28.25 & 0.851 & 0.202 \\
\midrule
ST-TAA (128) & 24.08 & 30.10 & 27.83 & \textbf{25.86} & 29.76 & 30.14 & \textbf{27.63} & 27.92 & \textbf{0.822} & \textbf{0.219} \\
Ours (16) & 23.58 & 28.80 & 26.90 & 24.69 & 28.40 & 29.34 & 26.54 & 26.89 & 0.741 & 0.330 \\
Ours (98) & \textbf{24.09} & \textbf{30.20} & \textbf{27.84} & 25.82 & \textbf{29.80} & \textbf{30.22} & 27.58 & \textbf{27.94} & \textbf{0.822} & 0.226 \\
Ours (128) & 24.13 & 30.32 & 27.91 & 25.90 & 29.91 & 30.30 & 27.65 & 28.02 & 0.829 & 0.217 \\
Ours (zero-shot, 16) & 23.66 & 28.96 & 26.99 & 24.83 & 28.56 & 29.33 & 26.67 & 27.00 & 0.746 & 0.332 \\
Ours (zero-shot, 128) & 24.14 & 30.33 & 27.92 & 25.92 & 29.93 & 30.29 & 27.66 & 28.03 & 0.829 & 0.217 \\
\bottomrule
\end{tabular}
\caption{Static quality on the MipNeRF360 held-out test views. \emph{2DGS} is the deterministic tile renderer of 2D Gaussian Splatting (Huang et al.). \emph{ST-TAA} is the baseline adapted from StochasticSplats (Kheradmand et al.): temporal accumulation over 128 static frames without a learned denoiser. \emph{Ref.\ (1024)} is the Monte-Carlo accumulation reference (main text, Experimental Setup). \emph{Ours} renders through the learned denoiser at 16, 98, or 128 accumulated frames; \emph{Ours (zero-shot)} deploys the room checkpoint on every scene with no per-scene denoiser training (main text, Zero-Shot Transfer). Per-scene PSNR with dataset averages of all metrics; bold marks the better of the equal-time pair, ST-TAA (128) vs.\ Ours (98), in each column (both when equal after rounding).}
\label{tab:static_mipnerf}
\end{table*}

\begin{table*}[t]
\centering
\footnotesize
\setlength{\tabcolsep}{2pt}
\begin{tabular}{l ccccccc ccc}
\toprule
 & \multicolumn{7}{c}{PSNR$\uparrow$ per scene} & \multicolumn{3}{c}{Average} \\
\cmidrule(lr){2-8} \cmidrule(lr){9-11}
Method & Barn & Caterpillar & Church & Courthouse & Ignatius & Meetingroom & Truck & PSNR$\uparrow$ & SSIM$\uparrow$ & LPIPS$\downarrow$ \\
\midrule
2DGS & 22.82 & 20.33 & 21.52 & 21.73 & 19.78 & 22.68 & 21.65 & 21.50 & 0.700 & 0.379 \\
Ref.\ (1024) & 22.39 & 19.84 & 19.87 & 21.17 & 19.11 & 21.42 & 21.04 & 20.69 & 0.675 & 0.388 \\
\midrule
ST-TAA (128) & \textbf{22.35} & \textbf{19.75} & \textbf{19.75} & \textbf{21.11} & \textbf{19.05} & 21.30 & 20.96 & \textbf{20.61} & \textbf{0.635} & \textbf{0.431} \\
Ours (16) & 22.07 & 19.46 & 19.42 & 20.83 & 18.79 & 21.01 & 20.72 & 20.33 & 0.548 & 0.496 \\
Ours (98) & 22.32 & 19.74 & \textbf{19.75} & 21.09 & 19.03 & \textbf{21.32} & \textbf{20.98} & 20.60 & 0.631 & 0.436 \\
Ours (128) & 22.34 & 19.77 & 19.78 & 21.11 & 19.05 & 21.34 & 21.00 & 20.63 & 0.640 & 0.428 \\
Ours (zero-shot, 16) & 22.04 & 19.40 & 19.37 & 20.80 & 18.74 & 20.97 & 20.71 & 20.29 & 0.539 & 0.501 \\
Ours (zero-shot, 128) & 22.34 & 19.76 & 19.77 & 21.10 & 19.05 & 21.34 & 20.99 & 20.62 & 0.639 & 0.428 \\
\bottomrule
\end{tabular}
\caption{Static quality on TanksAndTemples. All methods are trained on the full capture of each scene; we evaluate every 8th frame. Rows follow Table~\ref{tab:static_mipnerf}; bold marks the better of the equal-time pair, ST-TAA (128) vs.\ Ours (98), in each column (both when equal after rounding). \emph{Ours (zero-shot)} deploys the room checkpoint with no per-scene denoiser training (main text, Zero-Shot Transfer).}
\label{tab:static_tnt}
\end{table*}

\begin{table*}[t]
\centering
\small
\begin{tabular}{l ccccc ccccc}
\toprule
& \multicolumn{5}{c}{PSNR$\uparrow$ / LPIPS$\downarrow$} & \multicolumn{5}{c}{SSIM$\uparrow$ / CVVDP$\uparrow$} \\
\cmidrule(lr){2-6} \cmidrule(lr){7-11}
Track & 1\,spp & ST-TAA & SVGF & Ref.\ & Ours & 1\,spp & ST-TAA & SVGF & Ref.\ & Ours \\
\midrule
bicycle (161 frames) & 16.81 & 19.82 & 22.05 & 29.87 & \textbf{26.39} & 0.289 & 0.407 & 0.709 & 0.925 & \textbf{0.807} \\
                     & 0.639 & 0.588 & 0.346 & 0.088 & \textbf{0.284} & 4.28 & 5.29 & 5.23 & 8.34 & \textbf{6.84} \\
\addlinespace[2pt]
bonsai (166 frames)  & 19.28 & 23.00 & 28.17 & 34.54 & \textbf{31.98} & 0.313 & 0.486 & 0.868 & 0.967 & \textbf{0.914} \\
                     & 0.690 & 0.629 & 0.241 & 0.060 & 0.253 & 5.57 & 6.66 & 7.23 & 8.96 & \textbf{8.25} \\
\addlinespace[2pt]
counter (126 frames) & 18.05 & 20.97 & 23.98 & 33.76 & \textbf{30.35} & 0.257 & 0.367 & 0.766 & 0.956 & \textbf{0.880} \\
                     & 0.717 & 0.677 & 0.339 & 0.099 & \textbf{0.271} & 4.52 & 5.42 & 5.71 & 8.82 & \textbf{7.50} \\
\addlinespace[2pt]
garden (157 frames)  & 16.08 & 19.61 & 23.37 & 31.34 & \textbf{26.63} & 0.244 & 0.394 & 0.720 & 0.944 & \textbf{0.794} \\
                     & 0.606 & 0.531 & 0.308 & 0.067 & \textbf{0.238} & 4.60 & 5.81 & 6.11 & 8.80 & \textbf{7.15} \\
\addlinespace[2pt]
kitchen (181 frames) & 19.24 & 22.33 & 27.98 & 33.67 & \textbf{30.95} & 0.306 & 0.450 & 0.854 & 0.955 & \textbf{0.906} \\
                     & 0.643 & 0.579 & 0.240 & 0.066 & \textbf{0.192} & 5.69 & 6.58 & 7.40 & 8.93 & \textbf{8.07} \\
\addlinespace[2pt]
room (185 frames)    & 20.79 & 24.00 & 28.82 & 34.70 & \textbf{33.64} & 0.334 & 0.476 & 0.884 & 0.968 & \textbf{0.950} \\
                     & 0.697 & 0.647 & 0.232 & 0.089 & \textbf{0.171} & 5.46 & 6.52 & 7.24 & 8.80 & \textbf{8.29} \\
\addlinespace[2pt]
stump (179 frames)   & 17.54 & 20.61 & 24.59 & 32.89 & \textbf{28.66} & 0.229 & 0.347 & 0.692 & 0.936 & \textbf{0.815} \\
                     & 0.705 & 0.656 & 0.376 & 0.093 & \textbf{0.300} & 4.48 & 5.37 & 5.11 & 8.46 & \textbf{6.89} \\
\midrule
Average              & 18.26 & 21.48 & 25.57 & 32.96 & \textbf{29.80} & 0.282 & 0.418 & 0.785 & 0.950 & \textbf{0.867} \\
                     & 0.671 & 0.615 & 0.297 & 0.080 & \textbf{0.244} & 4.94 & 5.95 & 6.29 & 8.73 & \textbf{7.57} \\
\bottomrule
\end{tabular}
\caption{Track-averaged quality under free camera navigation on MipNeRF360, using the vanilla 2DGS tile renderer as the reference. Each track occupies two rows: the first reports PSNR (left) and SSIM (right), the second LPIPS (left) and CVVDP (right). Raw 1\,spp is the single-sample stochastic output; ST-TAA applies generic temporal accumulation without learned trust or spatial filtering; SVGF is the adapted variance-guided filtering baseline (Sec.~\ref{app:svgf}); Ref.\ (1024) is the per-frame Monte-Carlo ceiling (main text, Experimental Setup), not a practical method. The generic TAA warp fails to accumulate reliably under the heavy 1\,spp noise, while our learned trust gating keeps the temporal history usable.}
\label{tab:navigation_full}
\end{table*}

\begin{table*}[t]
\centering
\small
\setlength{\tabcolsep}{3.5pt}
\begin{tabular}{l cc cc cc cc}
\toprule
 & \multicolumn{2}{c}{PSNR$\uparrow$} & \multicolumn{2}{c}{SSIM$\uparrow$} & \multicolumn{2}{c}{LPIPS$\downarrow$} & \multicolumn{2}{c}{CVVDP$\uparrow$} \\
\cmidrule(lr){2-3} \cmidrule(lr){4-5} \cmidrule(lr){6-7} \cmidrule(lr){8-9}
Track & SS 1\,spp & SS 4\,spp & SS 1\,spp & SS 4\,spp & SS 1\,spp & SS 4\,spp & SS 1\,spp & SS 4\,spp \\
\midrule
bicycle & 15.70 & \textbf{20.08} & 0.275 & \textbf{0.483} & 0.634 & \textbf{0.523} & 3.65 & \textbf{6.18} \\
bonsai & 18.84 & \textbf{23.68} & 0.314 & \textbf{0.557} & 0.689 & \textbf{0.580} & 5.21 & \textbf{7.36} \\
counter & 18.13 & \textbf{22.60} & 0.273 & \textbf{0.475} & 0.702 & \textbf{0.597} & 4.31 & \textbf{6.47} \\
garden & 15.90 & \textbf{20.78} & 0.273 & \textbf{0.528} & 0.590 & \textbf{0.449} & 4.54 & \textbf{6.84} \\
kitchen & 18.83 & \textbf{24.48} & 0.313 & \textbf{0.594} & 0.641 & \textbf{0.488} & 5.62 & \textbf{7.71} \\
room & 21.05 & \textbf{25.50} & 0.365 & \textbf{0.605} & 0.668 & \textbf{0.560} & 5.35 & \textbf{7.21} \\
stump & 16.59 & \textbf{20.64} & 0.229 & \textbf{0.441} & 0.690 & \textbf{0.581} & 3.98 & \textbf{6.30} \\
\midrule
Average & 17.86 & \textbf{22.54} & 0.292 & \textbf{0.526} & 0.659 & \textbf{0.540} & 4.67 & \textbf{6.87} \\
\bottomrule
\end{tabular}
\caption{Per-track breakdown of the official StochasticSplats rows of the main text's navigation table: the official model on its own finetuned 3DGS scenes, measured against its own 1024\,spp accumulation (no ground truth exists on free camera tracks). SS 4\,spp renders through the official viewer's multi-sample path with a small fix so that each MSAA sample draws an independent stochastic sample (as released, all samples share one draw). All metrics are computed on every 10th track frame (the stride-10 dumps underlying the corresponding main-table rows; ColorVideoVDP uses the same stride-10 videos).}
\label{tab:ss_tracks}
\end{table*}

\begin{table*}[h]
\centering
\small
\setlength{\tabcolsep}{3.5pt}
\begin{tabular}{l cc cc cc cc}
\toprule
 & \multicolumn{2}{c}{PSNR$\uparrow$} & \multicolumn{2}{c}{SSIM$\uparrow$} & \multicolumn{2}{c}{LPIPS$\downarrow$} & \multicolumn{2}{c}{CVVDP$\uparrow$} \\
\cmidrule(lr){2-3} \cmidrule(lr){4-5} \cmidrule(lr){6-7} \cmidrule(lr){8-9}
Track & Ours & zero-shot & Ours & zero-shot & Ours & zero-shot & Ours & zero-shot \\
\midrule
bicycle & \textbf{26.39} & 25.86 & 0.807 & \textbf{0.817} & 0.284 & \textbf{0.255} & \textbf{6.84} & 6.56 \\
bonsai  & \textbf{31.98} & 30.91 & \textbf{0.914} & \textbf{0.914} & 0.253 & \textbf{0.205} & \textbf{8.25} & 7.83 \\
counter & \textbf{30.35} & 29.68 & 0.880 & \textbf{0.891} & 0.271 & \textbf{0.248} & \textbf{7.50} & 7.32 \\
garden  & 26.63 & \textbf{26.82} & 0.794 & \textbf{0.808} & 0.238 & \textbf{0.235} & \textbf{7.15} & 7.12 \\
kitchen & \textbf{30.95} & 30.61 & 0.906 & \textbf{0.914} & 0.192 & \textbf{0.162} & \textbf{8.07} & \textbf{8.07} \\
room    & \textbf{33.64} & 33.59 & \textbf{0.950} & 0.948 & \textbf{0.171} & 0.179 & \textbf{8.29} & 8.25 \\
stump   & \textbf{28.66} & 27.91 & 0.815 & \textbf{0.817} & 0.300 & \textbf{0.275} & \textbf{6.89} & 6.51 \\
\midrule
Average & \textbf{29.80} & 29.34 & 0.867 & \textbf{0.873} & 0.244 & \textbf{0.223} & \textbf{7.57} & 7.38 \\
\bottomrule
\end{tabular}
\caption{Per-track breakdown of the zero-shot transfer experiment (main text, Zero-Shot Transfer): the room checkpoint deployed verbatim on every scene, with the neural-view payload neutralised (the trust feature $V_2$ is set to zero and the spatial filter falls back to a fixed isotropic kernel, $x = y = 2$ and $\rho = 0$ in the covariance parameterisation of the main text). \emph{Ours} is the per-scene fine-tuned configuration of the main navigation table. The zero-shot variant trades a small PSNR/CVVDP margin for slightly better SSIM/LPIPS.}
\label{tab:zeroshot_tracks}
\end{table*}

\textbf{Denoiser stage.}
Table~\ref{tab:hyper} lists the full Stage-4 configuration.

\begin{table*}[h]
\centering
\small
\begin{tabular}{ll}
\toprule
Parameter & Value \\
\midrule
finetune steps & 15k (4k for fast transfer) \\
loss & L1 + $0.05$\,SSIM + $0.01$\,LPIPS \\
auxiliary vanilla-colour loss (accumulated path) & $0.2$ \\
temporal step ratio & $0.7$ \\
stochastic suffix length $K$ & $8$ \\
chain-length curriculum & $2 \to 128$ \\
max perturbation angle / position & $3.0^{\circ}$ / $0.003 \times$ scene radius \\
drift / lateral-dolly chain fractions & $0.25$ / $0.25$ \\
lateral dolly total length & $0.5$--$1.1$\,m \\
history caps $n_{\max}$ (denoised / accumulated) & $24$ / $128$ \\
near-static trust regularisers ($t_s$ / $t_h$) & $0.01$ / $0.01$ ($0.005$ / $0.005$ in fast transfer) \\
disocclusion mask: depth threshold $\tau_d$ & $0.02$ (relative) \\
disocclusion mask: min opacity / colour tolerance & $0.6$ / $0.05$ \\
disocclusion trust penalty weight & $0.03$ (fast transfer adds $0.05$ on both paths) \\
filter axis-symmetry regulariser & $0.01$ \\
gate warmup (forced denoised output) & $2000$ steps \\
gate scalar minimum scale & $10^{-4}$ \\
gate time constant $\tau$ & $8$ frames \\
\bottomrule
\end{tabular}
\caption{Stage-4 training configuration.}
\label{tab:hyper}
\end{table*}

\end{document}